\documentclass[journal]{IEEEtran}

\usepackage{times}
\usepackage{epsfig}
\usepackage{graphicx}
\usepackage{amsmath}
\usepackage{amssymb}
\usepackage{longtable}
\usepackage{amsfonts}
\usepackage{algorithmic}
\usepackage{algorithm}
\usepackage{color}
\usepackage{url}
\usepackage{tabularx}
\usepackage{array,multirow}
\usepackage{amssymb}

\begin{document}

\title{Sparse data interpolation using the geodesic distance affinity space}


\author{Mikhail~G.~Mozerov,~\IEEEmembership{~Member,~IEEE} and Fei Yang, 
     and Joost~van~de~Weijer,
\IEEEcompsocitemizethanks{\IEEEcompsocthanksitem M. Mozerov and F. Yang,  and J.van de Weijer are with the Computer Vision Center
of Department Informatics, Universitat Autònoma de Barcelona, Barcelona,
Spain, 08193.

F. Yang, is also with Key Laboratory of Information Fusion Technology,
Northwestern Polytechnical University, China
\protect\\
E-mail: mozerov@cvc.uab.es}
\thanks{}}


\markboth{IEEE SPL, ~Vol.~**, No.~**}%
{Shell \MakeLowercase{\textit{et al.}}: Sparse data interpolation using the geodesic distance affinity space}

\IEEEcompsoctitleabstractindextext{%
\begin{abstract}
In this paper, we adapt the geodesic  distance-based  recursive  filter~\cite{mozerov2017improved} to the sparse data interpolation problem.   The proposed technique is general and can be easily applied to any kind of sparse data.
We demonstrate the superiority over other interpolation techniques in three experiments for qualitative and quantitative evaluation. 
 In addition, we compare our method with the popular interpolation algorithm  presented in the EpicFlow  optical flow  paper~\cite{revaud2015epicflow}  that is intuitively motivated by a similar geodesic distance principle. 
The comparison shows that our algorithm is more accurate and considerably faster than the EpicFlow interpolation technique. 
\end{abstract}
\begin{IEEEkeywords}
Sparse data interpolation, geodesic distance filter, adaptive filter.
\end{IEEEkeywords}}
\maketitle
\IEEEdisplaynotcompsoctitleabstractindextext
\IEEEpeerreviewmaketitle
\section{Introduction}

Image scaling plays an important role in image processing. Especially when coarse-to-fine schemes are used. 
A major drawback of coarse-to-fine algorithms is error-propagation.  
For this reason, various interpolation methods were developed, 
from simple and popular algorithms of bilinear and bicubic interpolation to the wide class of polynomial and sinc-interpolations~\cite{yaroslavsky2013digital}. 

As a rule, scaling deteriorates the quality of the restored image due to various interpolation artefacts related to the loss of high-frequency information in the restored image. Difficulties of the sparse data interpolation problem increase severely when the data sparsity is irregular or possesses considerable gaps to be filled.  There are two main approaches to solve this specific variation of the sparse data interpolation task. One approach is to combine  a 
Delaunay triangulation~\cite{lee1980two}  and a barycentric Interpolation~\cite{berrut2004barycentric}.
Another method is the so called Nadaraya-Watson estimation~\cite{weber2008parallel}, 
where a desired value in any uncertain point is expressed by a sum of matches weighted by their proximity 
with a Gaussian kernel for a distance between the interpolated value point and its known neighbor value points. All these methods either fail to recover high-frequency information or in the image space lead to edge smoothing artefacts. 

Fortunately, if the sparse data has additional information correlated with the restored function we can solve the sparse data interpolation more accurately with the class of edge preserving filters.  The large class of edge-preserving smoothing filters~\cite{N1BF95,N2BF95,N3BF98} has received considerable attention in image processing, computer graphics, and computer vision. The filters have been applied to a wide variety of applications such as edit propagation~\cite{N6BF08,criminisi2010geodesic}, denoising~\cite{N7BF07,chen2013fast}, stereo matching~\cite{mei2011ADSNS},  optical flow~\cite{N8BF06}, video abstraction and demosaicing~\cite{N12BF03,N11BF06}. In general, the classic bilateral kernel is a function of the Euclidean distance in the joint color and spatial coordinates multidimensional space. The computational complexity of the brute-force implementation for this kind of kernels is highly demanding. Several fast algorithms were proposed in recent years~\cite{N21BF10,gastal2012adaptive,mozerov2015global,N14BF06}, where the approximation achieves high quality. The computational complexity for the fast realization usually depends on filter parameters that make this kind of filters less flexible and still demanding for several sets of parameters~\cite{N21BF10,gastal2012adaptive}. 

The topology of the  standard bilateral filter with Gaussian kernel does not fit well  to the stereo matching and optical flow estimation problems. This problem arises  because disparity or motion vector values of two near pixels can be considerably different but they might have similar color values in the cover image. In this case the bilateral filter would blur both output values.  
This ability to affect over color edges (e. g. the bilateral distance between two white pixels separated by a thin black line is considerably smaller than the geodesic distance) is useful for color based  segmentation, but can produce estimation artefacts for stereo and motion estimation. 
Therefore edge preserving filters based on the geodesic distance measure of a cover stereo or motion image  are more relevant to the above mentioned problems and also faster than the classic bilateral filters. We have to note that only a small part of image edges corresponds to motion boundaries, and this issue can cause inaccuracy for several non-confident regions, in particular for pixels which are isolated from pixels with known values.  Nevertheless,  the edge-preserving interpolation is still better than  the interpolation that does not use a cover image color information.      

This paper is partly inspired by the paper of Revaud  et al.~\cite{revaud2015epicflow}, where a sparse data interpolation method, called EpicFlow, was proposed. The main idea is to use geodesic distance to estimate the influence of known pixel values in a neighborhood of a recovered pixel value. The interpolation method was proposed and applied to optical flow estimation. The EpicFlow sparse data interpolation approach is used in several state-of-the-art  optical flow estimation algorithms (e.g. the DCflow~\cite{xu2017accurate} method). 

 However, this interpolation technique provides an heuristic problem solution that   consists of three heuristic algorithmic steps:  main edge extraction with  {"}structured edge detector{"}  (SED)~\cite{dollar2013structured}; Voronoi cells segmentation; and geodesic distance field approximation using Dijkstra{'}s algorithm~\cite{skiena1990dijkstra}. Each step includes its own set of parameters that are  weakly connected to each other. Despite the fact that the full pipeline of this interpolation is faster than direct geodesic distance implementation~\cite{liu2013joint} it is still computably demanding. A theoretical basis for edge-preserving filtering with geodesic distance has been proposed in~\cite{Gastal2011} and further extended in~\cite{Yang2012}. The latest approach~\cite{mozerov2017improved} improves the filter approximation in the sense of the geodesic distance based filtering accuracy and collateral artefacts suppression.

 Consequently, we   propose a simple and fast geodesic based interpolation using a bilateral filter with  geodesic distance kernel, based on the 
 method~\cite{mozerov2017improved}, where a fast and accurate approximation to the ideal filter with a geodesic kernel is proposed.  
 We have to note that the filter in~\cite{mozerov2017improved} was initially proposed for the denoising problem and we adopt the filter for the sparse data interpolation problem.  Finally, the proposed   approach faster, more general and with clearer theoretical motivation than the baseline algorithm~\cite{revaud2015epicflow}.  We  applied our interpolation method to the sparse optical flow data obtained by the DCflow~\cite{xu2017accurate} method and compared with the interpolation result of the interpolation in~\cite{revaud2015epicflow} on the same sparse data set. Formally  we included our interpolation method in the pipeline of the DCflow~\cite{xu2017accurate} method  and compared it with the  result of the same DCflow pipeline that included the EpicFlow interpolation (EFI) instead ours.   The comparison shows that our algorithm makes the fast version of the DCflow~\cite{xu2017accurate} pipeline more accurate than the EFI technique while being considerably faster.    

\section{Problem definition}\label{sec:def}
The interpolation problem can be defined via a more general solver for confidence mapping, because this strict definition makes the proposed algorithm clearly motivated.  In this case one aims to minimize the global mean squared  error between a known input data $y$ 
and a desired output solution $x$ as follows:
\begin{equation}
 \hat{x} = \mathop {\arg \min }\limits_x \sum\limits_{p \in {\cal V}} {\sum\limits_{q \in {\cal V}} {{w_{p,q}}{c_q}{{\left( {{x_p} - {y_q}} \right)}^2}} } 
\label{eq:Base_Problem}
\end{equation}
where $p,q \in {\cal V}$ corresponds to pixels or vertices and set   $ \left( {p,q} \right) \in {\cal E}$ to edges of an image graph $ {\cal G }= \left\{ { {\cal V}, {\cal E}} \right\}$.
A variable  $y_q$ is defined on all the known values of the input sparse data with non-zero confidence, and $x_p$ is the desired output function that has to be recovered. Usually confidence weights $c_q$  belong to the interval ${c_q} \in [0,1]$. In the case of the sparse data interpolation weights $c_q$ are exactly equaling 1 (known values) and 0 elsewhere, or formally: 
\[{c_q} = \left\{ \begin{array}{l}
 \begin{array}{*{20}{c}}
   1 & {{\rm{if}}} & {{y_q}} & {\rm{is}} & {{\rm{known}}}  \\
\end{array} \\ 
 \begin{array}{*{20}{c}}
   0 & {{\rm{elsewhere.}}}  \\
\end{array} \\ 
 \end{array} \right.\]
The weights $w_{p,q}$ in Eq.~(\ref{eq:Base_Problem}) define the influence of a known input value $y_q$ on the desired output value $x_p$. For conventional interpolation, where the only known information is the values $y_q$, this influence usually  depends only on the Euclidean or barycentric distance between pixels $q$ and $p$ in the image plane. However in the case of the optical flow or stereo matching  we can use the cover image, whose values usually strongly correlate with the optical flow or the disparity map values. It is reasonable  to define the bilateral affinity space were the distance depends also on the cover image values $I_q$ and $I_p$. 

Consequently, we define the sparse data interpolation problem via the general functional minimization 
(Eq.~(\ref{eq:Base_Problem})).
We extend the sparse data set $y$ to the full image graph domain ${\cal V}$ by combination with confidence factors in the form:
\[{{\tilde y}_p} = \left\{ \begin{array}{l}
 \begin{array}{*{20}{c}}
   {{y_q}} & {{\rm{if}}} & {{y_q}}& {{is}} & {{\rm{known}}}  \\
\end{array} \\ 
 \begin{array}{*{20}{c}}
   0 & {{\rm{elsewhere.}}}  \\
\end{array} \\ 
 \end{array} \right. \]
It follows then the solution in Eq.~(\ref{eq:Base_Problem}) can be  obtained in the closed-form as a fraction of two  standard non-normalized bilateral filters:
\begin{equation}
\hat{x}_p = \frac{{\sum\limits_{q \in {\cal V}} {{w_{p,q}}{{\tilde y}_q}} }}{{\sum\limits_{q \in {\cal V}} {{w_{p,q}}{c_q}} }}.
\label{eq:Base_blf}
\end{equation}
Finally,  the formula in Eq.~(\ref{eq:Base_blf}) represented as the bilateral filter  is the desired closed-form solution for the general sparse data interpolation problem  Eq.~(\ref{eq:Base_Problem}).   

Application of the classical filters has several drawbacks, and
we propose an interpolation method that is based on the geodesic distance affinity space.  

 The geodesic distance is a generalization of the straight line distance in the Euclidean space to the distance measure in a curved space. In the case of images, the geodesic distance is defined as the shortest path on the surface between two points. Here the surface is formed by the image value function defined on the 2D spatial domain.

For the geodesic distance based filter weights $w_{p,q}$ are usually chosen  (e. g. ~\cite{Gastal2011}) as follows
\begin{equation}
{w_{p,q}} = {e^{ - a{d_{p,q}}}}
 \label{eq:pp}
 \end{equation}
which makes the filter recursive. Thus the weight  $ e^{ - a{d_{p,q}}} $  defines a geodesic distance based affinity between any two image pixels, and the variable  ${d_{p,q}}$  is the geodesic distance between image pixels $\left( {p,q} \right)$ which for an image ${I_p}$ can be defined on the discrete grid graph as 
\begin{equation}
\begin{array}{l}
 {d_{p,q}} = \mathop {\min }\limits_{{P_{p,q}}} \sum\limits_{\varepsilon  \in {P_{p,q}}} {{u_\varepsilon }} , \\
 {u_{\varepsilon  = \left( {k,l} \right)}} = \left\| {{I_k} - {I_l}} \right\| + \delta , \\
 \end{array}
\label{eq:GD_def}
\end{equation}
where ${P_{q,p}}$ is a path between two graph vertices $\left( {p,q} \right)$ and $\delta $ is the spatial interval related to discretization. 
Note that the  parameters $a$ and $\delta $  in Eqs.~(\ref{eq:pp},\ref{eq:GD_def}) 
approximately correspond to the parameters of the classic bilateral filter with the Gaussian kernel as follows
\begin{equation}
a = \frac{2}{{\sigma _r^2}},{\rm{ }}\delta  = \frac{{\sigma _r^2}}{{\sigma _s^2}}.
 \label{eq:K_par}
\end{equation}
From Eq.~(\ref{eq:GD_def}) one can derive the following  useful relation
\begin{equation}
{w_{p,q}}  =  {e^{ - a\mathop {\min }\limits_{P_{q,p}} \sum\limits_{\varepsilon  \in P_{p,q}} {{u_\varepsilon }} }}   =  \mathop {\max }\limits_{P_{p,q}} \prod\limits_{\varepsilon  \in {P_{p,q}}} {{
e^{-a{u_\varepsilon }}}}. 
\label{eq:PR_def}
\end{equation}

\begin{figure}[t]
\centering
\includegraphics[width=0.46\textwidth]{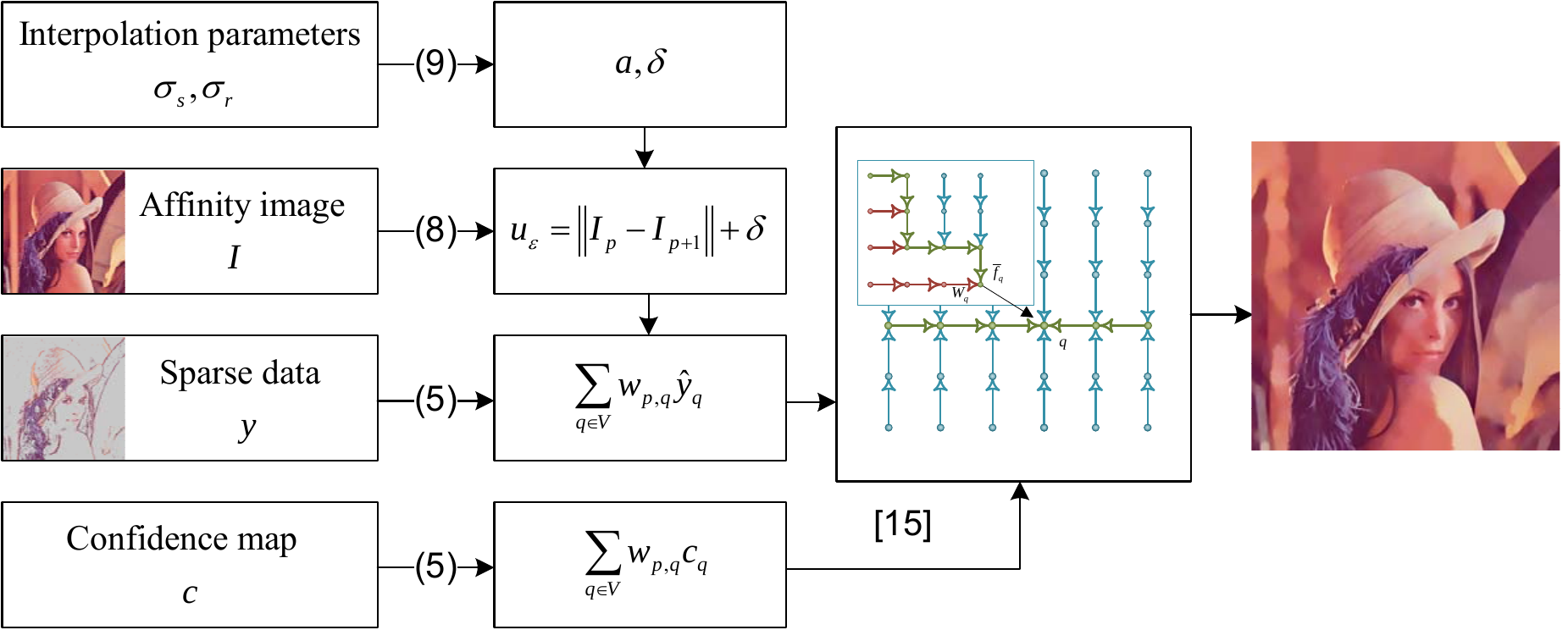}
\caption {The pipeline of the proposed interpolation method.The final step includes fast sums calculation with recursive calculation trees~\cite{mozerov2017improved}.}
\label{fg:scheme}
\end{figure}

In~\cite{mozerov2017improved} it is shown that both filter sums in Eq.~(\ref{eq:Base_blf}) can be calculated  recursively using specific calculation trees (in the paper called the optimal tree).
The full calculation tree, which corresponds to this fast algorithm~\cite{mozerov2017improved}, is composed by four quadrant-domains (or branches of the tree). One of four branches is illustrated inside our algorithm scheme in Fig.~\ref{fg:scheme},
 and $W_q$ represents the sum of weights  $w_{p,q}$   for the first quadrant, when ${{\tilde f}_q}$ gives the numerator sum in Eq.~(\ref{eq:Base_blf})  also for this quadrant. 
 The pipeline of our algorithm is shown in Fig.~\ref{fg:scheme}.
\section{Experiments}\label{sec:experiments}
 \begin{table*}
 \centering
  \caption{ Results comparison: the DCflow pipeline that includes our method and the DCflow pipeline that includes the EFI approach with the  Endpoint  error  metric for every MPI training set sequence.  
  The top part of this table is based on the non-occluded pixels,  and the middle on all pixels. 
  In the bottom part of the table the computation time per frame in seconds is given.
 }
 \tiny
 \begin{tabular}
 {c p{0.25cm}p{0.25cm}p{0.25cm}p{0.25cm}
p{0.25cm} p{0.25cm} p{0.25cm}p{0.25cm}p{0.25cm}p{0.25cm}
 p{0.25cm}p{0.25cm} p{0.25cm}p{0.25cm}
 p{0.25cm}p{0.25cm}p{0.25cm}p{0.25cm}p{0.25cm}
 p{0.25cm}p{0.25cm}p{0.25cm}p{0.25cm}p{0.25cm}}

 \hline
 &
 \parbox[b]{0.5mm}{\multirow{1}{*}
 {\rotatebox[origin=c]{90}{average~~~}}}&
 \parbox[b]{0.5mm}{\multirow{1}{*}
 {\rotatebox[origin=c]{90}{alley-1~~~}}}&
 \parbox[t]{0.5mm}{\multirow{1}{*}
 {\rotatebox[origin=c]{90}{alley-2~~~}}}& 
 \parbox[t]{0.5mm}{\multirow{1}{*}
 {\rotatebox[origin=c]{90}{ambush-2~~~}}}&
 \parbox[t]{0.5mm}{\multirow{1}{*}
 {\rotatebox[origin=c]{90}{ambush-4~~~}}}&
 \parbox[t]{0.5mm}{\multirow{1}{*}
 {\rotatebox[origin=c]{90}{ambush-5~~~}}}&
 \parbox[t]{0.5mm}{\multirow{1}{*}
 {\rotatebox[origin=c]{90}{ambush-6~~~}}}&
 \parbox[t]{0.5mm}{\multirow{1}{*}
 {\rotatebox[origin=c]{90}{ambush-7~~~}}}&
  \parbox[t]{0.5mm}{\multirow{1}{*}
 {\rotatebox[origin=c]{90}{bamboo-1~~~}}}&
 \parbox[t]{0.5mm}{\multirow{1}{*}
 {\rotatebox[origin=c]{90}{bamboo-2~~~}}}&
  \parbox[t]{0.5mm}{\multirow{1}{*}
  {\rotatebox[origin=c]{90}{bandage-1~~~}}}&
  \parbox[t]{0.5mm}{\multirow{1}{*}
  {\rotatebox[origin=c]{90}{bandage-2~~~}}}&
  \parbox[t]{0.5mm}{\multirow{1}{*}
  {\rotatebox[origin=c]{90}{cave-2~~~}}}&
  \parbox[t]{0.5mm}{\multirow{1}{*}
  {\rotatebox[origin=c]{90}{cave-4~~~}}}&
  \parbox[t]{0.5mm}{\multirow{1}{*}
  {\rotatebox[origin=c]{90}{market-2~~~}}}&
  \parbox[t]{0.5mm}{\multirow{1}{*}
  {\rotatebox[origin=c]{90}{market-5~~~}}}&
  \parbox[t]{0.5mm}{\multirow{1}{*}
  {\rotatebox[origin=c]{90}{market-6~~~}}}&
  \parbox[t]{0.5mm}{\multirow{1}{*}
  {\rotatebox[origin=c]{90}{mountain-1~~~}}}&
  \parbox[t]{0.5mm}{\multirow{1}{*}
  {\rotatebox[origin=c]{90}{shaman-2~~~}}}&
  \parbox[t]{0.5mm}{\multirow{1}{*}
  {\rotatebox[origin=c]{90}{shaman-3~~~}}}&
  \parbox[t]{0.5mm}{\multirow{1}{*}
  {\rotatebox[origin=c]{90}{sleeping-1~~~}}}&
  \parbox[t]{0.5mm}{\multirow{1}{*}
  {\rotatebox[origin=c]{90}{sleeping-2~~~}}}&
  \parbox[t]{0.5mm}{\multirow{1}{*}
  {\rotatebox[origin=c]{90}{temple-2~~~}}}&
  \parbox[t]{0.5mm}{\multirow{1}{*}
  {\rotatebox[origin=c]{90}{temple-3~~~}}}
  \\
  &&&&&&&&&&&&&&&&&&&&&&&&\\
  Method&&&&&&&&&&&&&&&&&&&&&&&&\\
  &&&&&&&&&&&&&&&&&&&&&&&&\\
   &&&&&&&&&&&&&&&&&&&&&&&&\\
\hline
  DCF + Ours (nocc) &\bf{6.91}&\bf{1.35}&\bf{1.47}&\bf{28.0}&\bf{28.3}&9.74&\bf{15.4}&1.68&\bf{1.75}
  &\bf{2.89}&\bf{1.46}&\bf{1.47}&12.7&\bf{9.87}&2.28&\bf{14.1}&{5.68}&1.59&\bf{0.82}
  &\bf{1.18}&\bf{0.92}&\bf{1.09}&\bf{5.33}&\bf{9.77}\\
 DCF + EFI& 7.25 &1.38&1.53&29.3&29.1&\bf{9.42}&18.7&\bf{1.46}&1.80&2.92&1.59&1.55&\bf{12.5}
 &9.97&\bf{2.25}&16.0&\bf{4.82}&\bf{1.50}&0.95&1.48&1.22&1.16&5.36&10.6\\[0.07cm]
 \hline
   DCF + Ours (all) &\bf{9.86}&\bf{1.60}&\bf{1.68}&\bf{37.0}&32.9&16.5&\bf{19.5}&2.75&\bf{1.95}&4.66&\bf{2.09}&\bf{1.64}
&\bf{18.2}&\bf{12.4}&\bf{3.14}&\bf{27.4}&11.3&2.25&\bf{0.87}&\bf{1.27}&\bf{0.92}&\bf{1.09}&\bf{8.77}&\bf{17.0}
   \\
 DCF + EFI& 9.95 &1.64&1.75&38.0&\bf{32.0}&\bf{15.5}&21.6&\bf{2.40}&2.01&\bf{4.31}&2.29&1.78&18.7
 &12.6&3.21&29.3&\bf{8.08}&\bf{1.95}&0.99&1.59&1.22&1.16&9.84&17.1\\[0.07cm]
 \hline
  \hline

  Time in seconds \\
  \hline
    Ours &\bf{0.32}&\bf{0.32}&\bf{0.32}&\bf{0.32}&\bf{0.32}&\bf{0.32}&\bf{0.32}&\bf{0.32}&\bf{0.32}&\bf{0.32}&\bf{0.32}&\bf{0.32}&\bf{0.32}&\bf{0.32}&\bf{0.32}&\bf{0.32}
   &\bf{0.32}&\bf{0.32}&\bf{0.32}&\bf{0.32}&\bf{0.32}&\bf{0.32}&\bf{0.32}&\bf{0.32}\\
  EFI& 2.71&3.53&3.49&1.20&1.49&2.15&1.56&2.37&3.39&3.33&3.22
 &3.34&2.43&2.70&2.27&1.97&2.51&1.86&3.47&3.46&3.50&3.71
 &2.73&1.64\\[0.07cm]
 \hline
 \rule{0cm}{0.5cm}
 \end{tabular}

\label{table:err}
 \end{table*}

The experiments have been designed to demonstrate the potential of  the proposed approach. They are divided into two parts:  
First, several experiments  are performed to give a qualitative and quantitative comparison between the proposed edge preserved interpolation and a traditional interpolation method.  Also we compare our algorithm with the bilateral based interpolation technique.
Second, we analyze the advantage of our approach over the EFI method~\cite{revaud2015epicflow}  for the optical flow problem. 
In our experiments, we used the MPI Sintel data set~\cite{Butler:ECCV:2012} and the fast version of  the DCflow~\cite{xu2017accurate} method pipeline.


For the  first, second and the third  experiments, we use two example stereo images and corresponding ground truth disparity maps: the Motorcycle and the Art with the disparity scopes  1-140 and 1-128 respectively. 
Featureless image regions are a problem for stereo matching and optical flow computation. and feature-based sampling could lead to more accurate results. In contrast, the regular downsampling is more simple and popular. For our first three model experiments we propose three different downsampling methods:   a pure feature-based sampling; a trade-off downsampling algorithm that includes both feature-based and regular downsampling techniques; and a pure regular-grid downsampling. 

\begin{figure}[t]
\begin{center}
\small

\begin{tabular}{c}
  Sparse~data~~~~~~~~~~[23]~~~~~~~~Bilateral kernel~~~~~~~Ours~~~~~~~\\ 
\includegraphics[width=0.94\columnwidth]{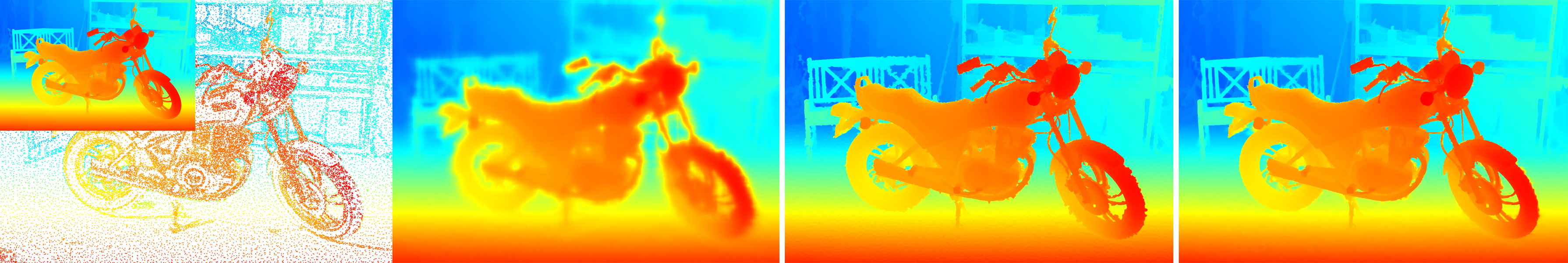}  \\
 ~~~~~(a)~~~~~~~~~~(b)~~~6.25~~~~~~~(c)~~~3.77 ~~~~~~~~(d)~~~3.31 \\
 \includegraphics[width=0.94\columnwidth]{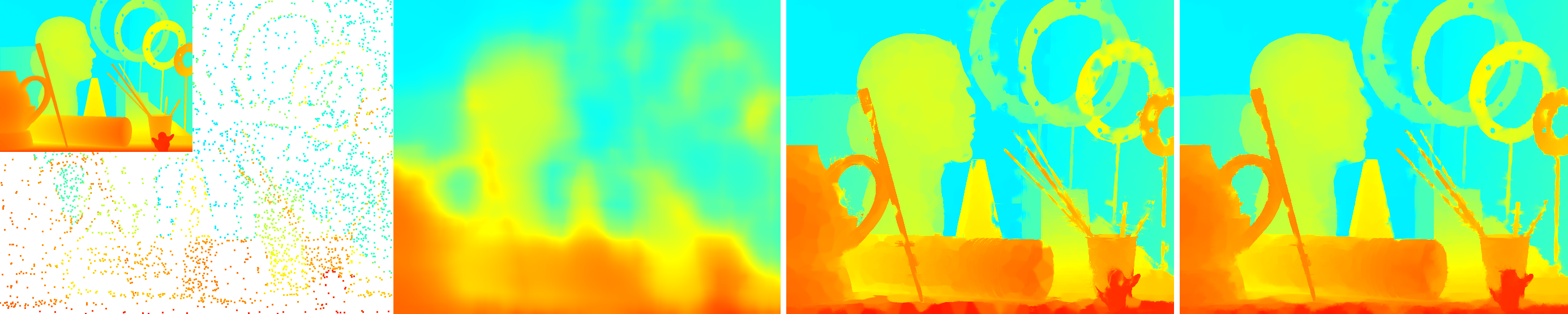}  \\
 ~~~~~(e)~~~~~~~~~~(f)~~~8.39~~~~~~~(g)~~~4.47~~~~~~~~(i)~~~3.78 \\
\end{tabular}
\caption{ Visual and quantitative comparison of the different interpolation methods for the non-uniform data sparsity distribution (first experiment).
The first and the second rows represents the sparse data and relevant interpolation results with the density equal to 4\% and  1\% respectively. The root-mean-square error is given below the image.} 
\label{fg:car}
\end{center}
\end{figure}

\begin{figure}[t]
\small
\begin{center}

\begin{tabular}{c}
Sparse~data~~~~~~~~~~[23]~~~~~~~~Bilateral kernel~~~~~~~Ours~~~~~~~\\
\includegraphics[width=0.94\columnwidth]{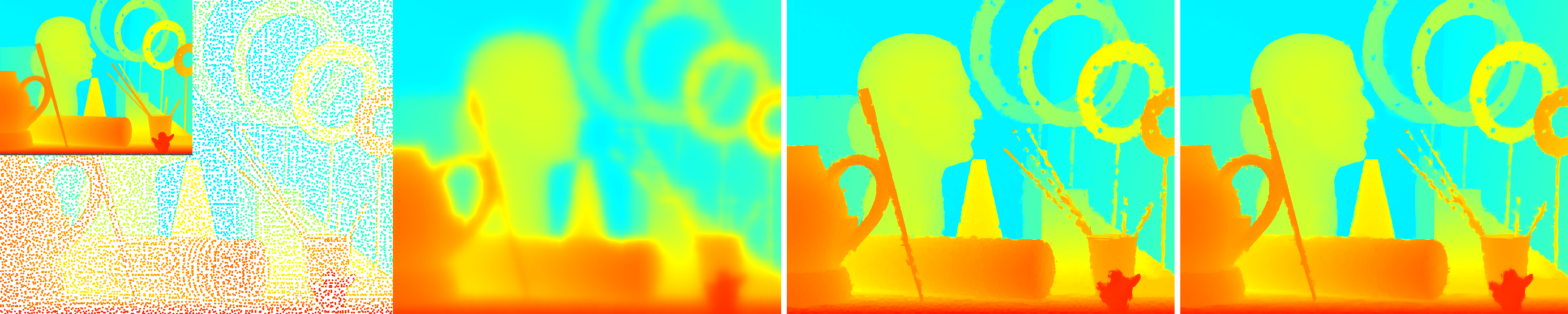}  \\
 ~~~~~(a)~~~~~~~~~~(b)~~~5.59~~~~~~~(c)~~~3.58~~~~~~~~(d)~~~3.30\\
 \includegraphics[width=0.94\columnwidth]{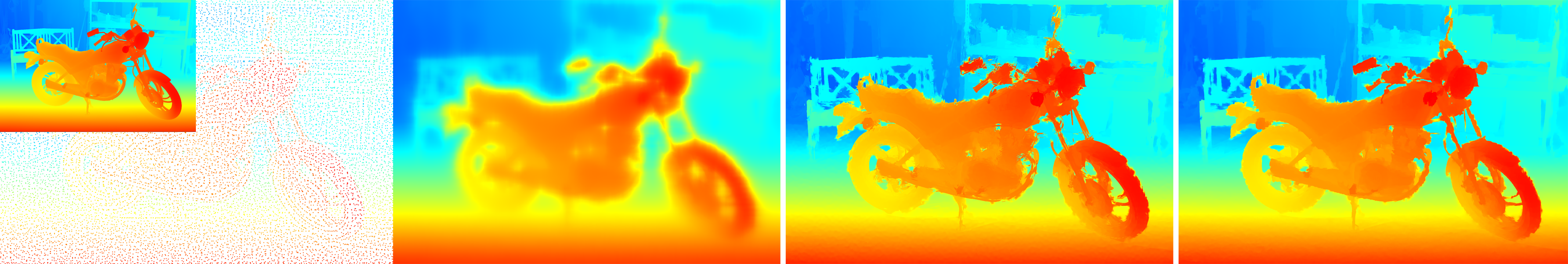}  \\
 ~~~~~(e)~~~~~~~~~~~(f)~~~7.99~~~~~~~(g)~~~7.41~~~~~~~~(i)~~~6.45\\
\end{tabular}
\caption{  Visual and quantitative comparison of the different interpolation methods for the uniform data sparsity distribution (second experiment).
The first and the second rows represents the sparse data and relevant interpolation results with the density equals to 4\% and  1\% respectively. The root-mean-square error is given below the image. }  
\label{fg:Lena}
\end{center}
\end{figure}

The first sparse data set is obtained by downsampling disparity values of a disparity map mostly near  visual edges of the image. In other words, we label a pixel $p$ as known if the norm of the  image gradient is $ \left\| {\vec \nabla {I_p}} \right\| > T $ and as  unknown elsewhere. The threshold $T$ is chosen in such a way that the density of known pixels is equal to a chosen constant. 
The second data set for our model experiment is obtained by downsampling pixels values of a disparity map almost uniformly. We partition a full original image into equal squared patches and label a pixel $p$ as known if the gradient value of the corresponding image pixel reaches maximum values inside the current patch, and as an unknown elsewhere.  
The sparse data for both experiments obtained using Motorcycle and Art and the relevant ground truth  disparity maps are illustrated in  Fig.~\ref{fg:car}-~\ref{fg:Lena}(a) and (e) with the density equal to 4\% for (a) and 1\% for (e). 
Then the sparse data is interpolated by the proposed  edge preserved algorithm, using the corresponding stereo  images for the affinity space calculation. We also interpolate the same sparse data with the Nadaraya-Watson estimation~\cite{weber2008parallel} and the classic bilateral filter. The quantitative evaluation is represented under each interpolated disparity map, where interpolation errors relative to ground truth is measured by the root-mean-squared error (RMSE) metric. The density of the known pixels in the first data set is distributed quite non-uniformly due to the image edge distribution. Thus the standard methods (for example Nadaraya-Watson estimation) loses almost all edge information after interpolation: Fig.~\ref{fg:car}(b) and (f). In contrast, our method and interpolation with the bilateral kernel  keep the meaningful edges as  it is illustrated in Fig.~\ref{fg:car}(d),(i), (c), (g) respectively. However, our method is more accurate than the bilateral kernel interpolation. Also our method does not possess artefacts due to color communication  over image edges as in the case of the bilateral interpolation.
For the second experiment the qualitative evaluation shows that the  Nadaraya-Watson estimation~\cite{weber2008parallel} in this case demonstrates better visual performance and does not loose essential information as we observed 
the first experiment shown in Fig.~\ref{fg:Lena}(b) and (f)
is still considerably worse than the edge preserved approach.   Our method and interpolation with the bilateral kernel keep the base edges  as  illustrated in Fig.~\ref{fg:Lena}(d),(i), (c), (g) respectively. 

\begin{figure}[t]
\centering
\includegraphics[width=0.43\textwidth]{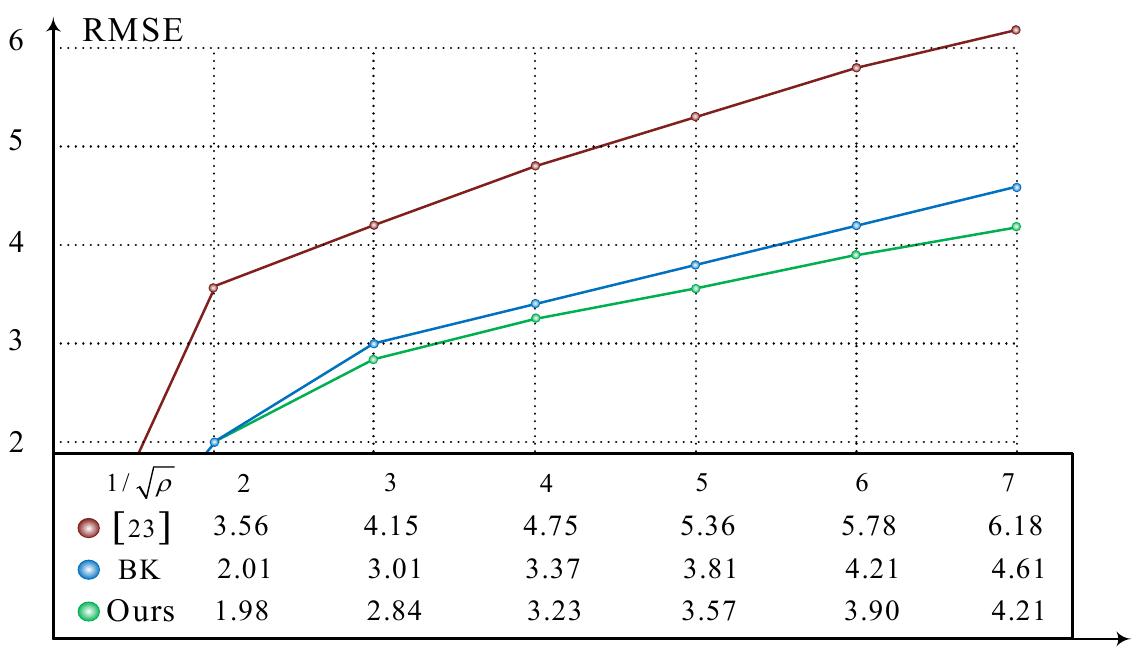}
\caption { Quantitative evaluation  of the interpolation  with respect to the inverse-root-density of known pixels  for all three considered interpolation methods: Nadaraya-Watson~\cite{weber2008parallel}, with the bilateral kernel (BK) and ours.} 
\label{fg:exp}
\end{figure}

The third experiment uses  known pixels of the Motorcycle disparity map that spread regularly (with a fixed sampling step) and evaluates accuracy of the interpolation with respect to the density $\rho$ of known pixels. The results are illustrated in   Fig.~\ref{fg:exp}, where the domain variable $1/\sqrt \rho$ is equal to the sampling step. One can see that the quality of interpolation is almost linear with respect to the inverse-root-density for all three considered interpolation methods: Nadaraya-Watson~\cite{weber2008parallel}, with the bilateral kernel (BK) and the proposed.  


\begin{figure}[t]
\begin{center}
\begin{tabular}{cc}

\includegraphics[width=0.25\columnwidth]{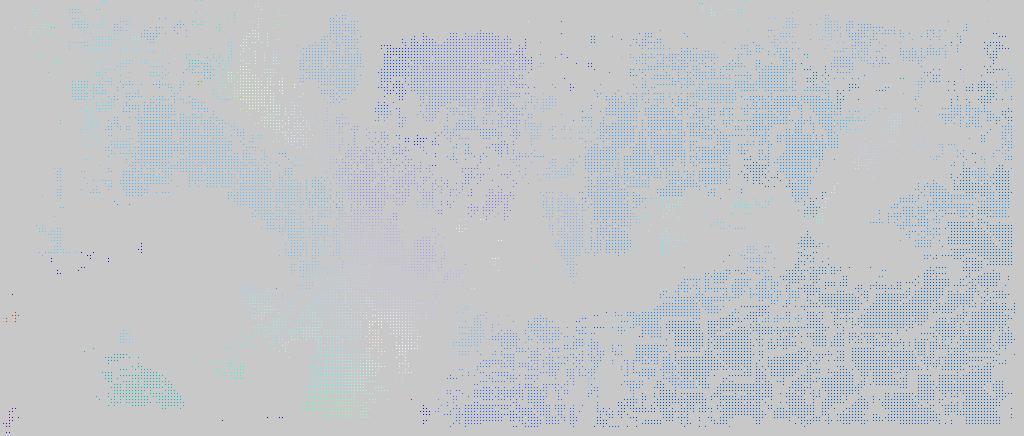} 
~~~~~~~~~~~&~~~~~~~~~~
\includegraphics[width=0.25\columnwidth]{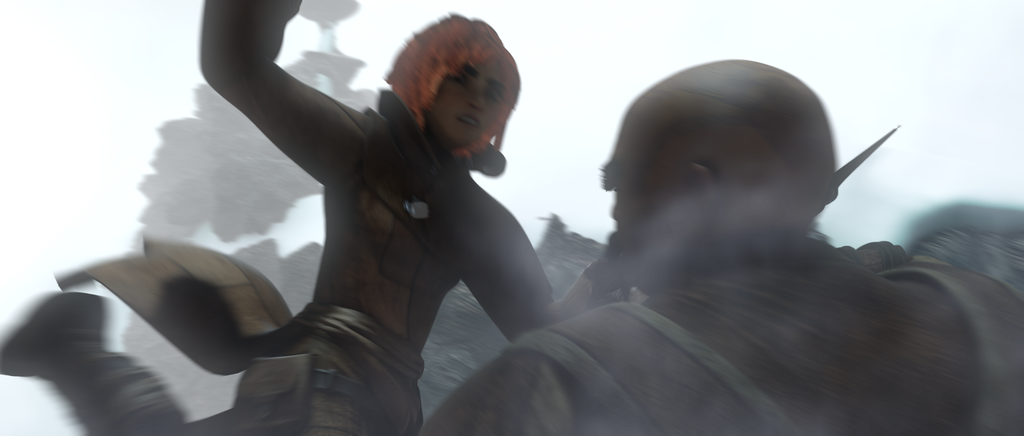}  \\
\end{tabular}
\begin{tabular}{ccc}
(a) & ~~ & (b) \\
\includegraphics[width=0.25\columnwidth]{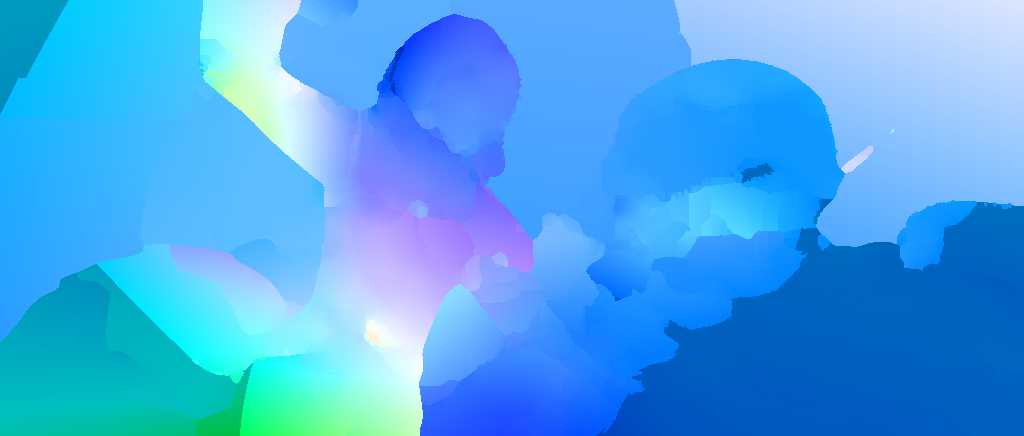} &
\includegraphics[width=0.25\columnwidth]{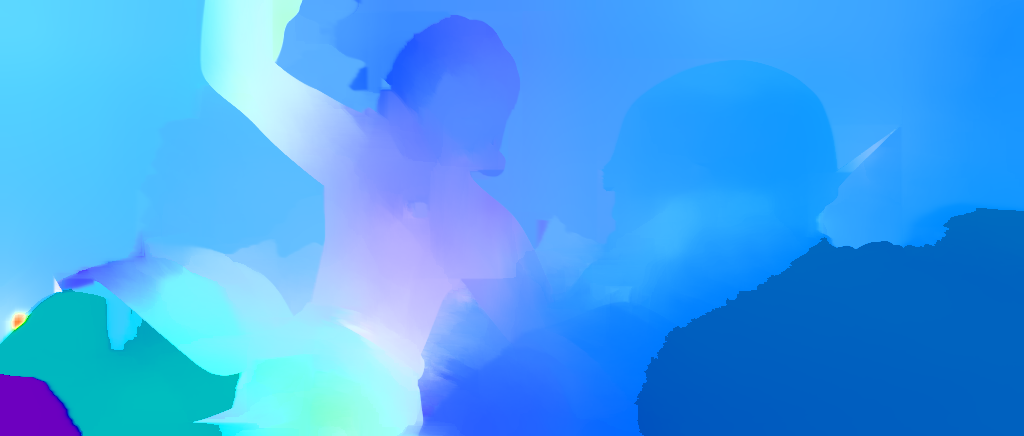} &
\includegraphics[width=0.25\columnwidth]{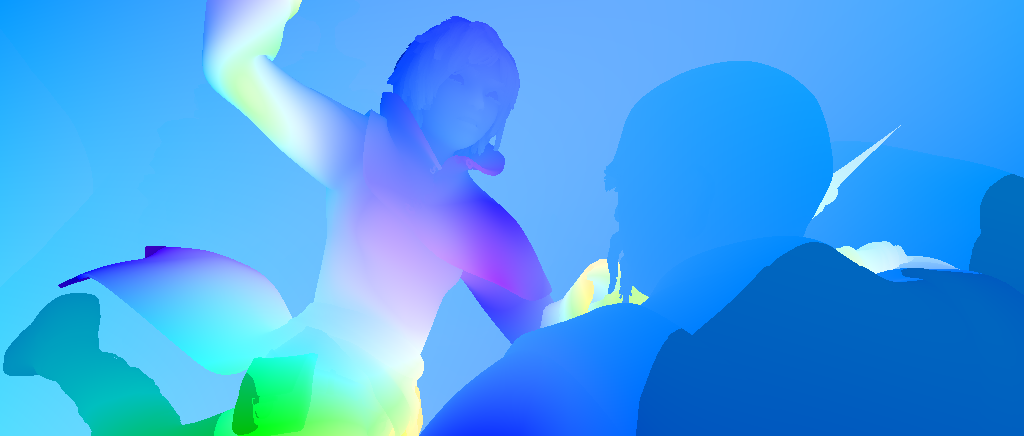} \\
(c) & (d)  & (e)\\
\end{tabular}
\caption{ 
Visual comparison of the proposed edge preserving interpolation
performance for the Ambush test image versus the EFI method:
(a) - the sparse data obtained by the DCFlow method of the MPI Ambush-6 sequence 12th frame; 
(b) - the relevant 12th frame image; 
(c) - the result of interpolation with the EFI method;
(d)  - the result of interpolation with  the proposed algorithm; 
(e) - the ground truth;}

\label{fg:ambush}
\end{center}
\end{figure}



The next experiment demonstrates the advantage of using the proposed geodesic distance based interpolation  applied to optical flow estimation in comparison with  the popular  interpolation method~\cite{revaud2015epicflow}.  The sparse data set for this experiment is  the result of optical flow estimation obtained by another state-of-the-art algorithm~\cite{xu2017accurate} using the MPI Sintel training data set~\cite{Butler:ECCV:2012}. Formally  we include our interpolation method in the pipeline of the DCflow~\cite{xu2017accurate} method  and compare it with the  result of the same DCflow pipeline that included the EFI instead ours.  The MPI training data set includes 23 different video sequences, in turn each sequence consists of up to 50 frames with the known ground truth optical flow. 
Fig.~\ref{fg:ambush} illustrates result of interpolation of  sparse data Fig.~\ref{fg:ambush}(a) with two methods: Fig.~\ref{fg:ambush}(c) - EFI~\cite{revaud2015epicflow}; Fig.~\ref{fg:ambush}(d) - the proposed algorithm; Fig.~\ref{fg:ambush}(e) - the ground truth result. Fig.~\ref{fg:ambush}(b) - illustrates an affinity image or a current frame   that correspond to the estimated motion vectors map. We can see that the EFI technique produces false segmentation in the image scene background, thus decreasing accuracy of the interpolation step.Note that our method provides smoother results.  

Quantitative comparison of our method and  EFI is illustrated in Table~\ref{table:err}, where comparison is given for every sequence of the MPI training data set. Our method provides quantitatively better results than the EFI in interpolation accuracy for the  endpoint  error metric over the average of all sequences as 9.86 : 9.95 (all pixels mask), and 6.91 : 7.25 (non-occluded pixels mask). Here for our algorithm we set parameters $\sigma _r=50$ and $ \sigma _s=100$ for all sequences.

Our algorithm is considerably faster than the EFI approach and this fact is illustrated in Table~\ref{table:err} (bottom part). From the table one can see that the computational time per frame of  the EFI depends on the considered sequence. The reason is that the computational time of the EFI directly depends on the sparsity (or density of the known pixels per image). We found that this time is equal to 3.2 sec for one frame with size 1024x436 for 1/9 sparsity factor, 0.5 sec for 1/100 sparsity factor, and 0.3 for 1/1000 sparsity factor.    
In contrast,  our method belongs to the  O(1) class of algorithms and the computational time per frame depends only on the size of the processed image. 

\section{Conclusions}\label{sec:conclusions}
In this work we  developed a fast and flexible sparse data interpolation algorithm using the geodesic  distance affinity filter~\cite{mozerov2017improved} and apply the derived pipeline to the optical flow estimation problem. Moreover, we found that our approach is more general, faster and with clearer theoretical motivation in comparison with the EFI approach.
\section*{Acknowledgements}
This work has been supported by the Spanish project TIN2015-65464-R, TIN2016-79717-R  (MINECO/FEDER).
and the COST Action IC1307 iV\&L Net (European Network on Integrating Vision and Language), supported by COST (European Cooperation in Science and Technology). 

{\small
\bibliographystyle{ieee}
\bibliography{egbib}
}

\end{document}